\documentclass[12pt,twoside]{article} 
 
 \usepackage[pdftex]{graphicx}
\usepackage{algorithm}
\usepackage{algorithmic}
\usepackage[squaren,Gray]{SIunits}
\date{} 
\begin{document} 
\centerline{\bf  International Journal of Engineering and Technology (IJET) VOL:4(°5).P. 268--276. 2012} 
\centerline{} 
\centerline{} 
\centerline {\Large{\bf  Application of Symmetric Uncertainty}} 
\centerline{} 
\centerline{\Large{\bf   and Mutual Information to Dimensionality   }} 
\centerline{}
\centerline{\Large{\bf Reduction of  and Classification}} 
\centerline{} 
\centerline{\Large{\bf  Hyperspectral Images}} 
\centerline{} 
\centerline{\bf {ELkebir Sarhrouni*, Ahmed Hammouch** and Driss Aboutajdine*}} 
\centerline{} 
\centerline{*LRIT,Faculty of Sciences, Mohamed V - Agdal University, Morocco} 
\centerline{**LRGE,ENSET, Mohamed V - Souissi University, Morocco} 

\centerline{sarhrouni436@yahoo.fr, hammouch\_a@yahoo.com, aboutaj@fsr.ac.ma}

\centerline{}

\centerline{} 
\newtheorem{Theorem}{\quad Theorem}[section] 
\newtheorem{Definition}[Theorem]{\quad Definition} 
\newtheorem{Corollary}[Theorem]{\quad Corollary} 
\newtheorem{Lemma}[Theorem]{\quad Lemma} 
\newtheorem{Example}[Theorem]{\quad Example}

\begin{abstract}
Remote sensing is a technology to acquire data  for disatant substances, necessary to construct a model knowledge for applications as classification. Recently Hyperspectral Images (HSI)  becomes a high technical tool that the main goal is to classify the point of a region. The HIS is more than a hundred bidirectional measures, called bands (or simply images), of the same region called Ground Truth Map (GT). But some bands are not relevant because they are affected by different atmospheric effects; others contain redundant information; and  high dimensionality of HSI features make the accuracy of classification lower. All these bands can be important for some applications; but for the classification a small subset of these is relevant. The problematic related to HSI is the dimensionality reduction. Many studies use mutual information (MI) to select the relevant bands. Others studies use the MI normalized forms, like Symmetric Uncertainty,  in medical imagery applications. In this paper we introduce an algorithm based also on MI to select relevant bands and it apply the Symmetric Uncertainty coefficient to control redundancy and increase the accuracy of classification. This algorithm is feature selection tool and a Filter strategy. We establish this study on HSI AVIRIS 92AV3C. This is an effectiveness, and fast scheme to control redundancy.
\end{abstract}



%

\section{Introduction}
In the feature classification domain, the choice of data affects widely the results. The problem of feature selection is commonly reencountered when we have $N$ features (or attributes) that express $N$ vectors of measures for $C$ substances (called classes). The problematic is to find $K$ vectors among $N$, such as relevant and no redundant ones; in order to classify substances. The number of selected vectors $K$  must be lower than $N$ because when $N$ is so large that needs many cases to detect the relationship between the vectors and the classes (Hughes phenomenon) [10] . No redundant features (vectors) because they complicate the learning system and product incorrect prediction [14]. Relevant vectors means there ability to predicate the classes. The Hyperspectral image (HIS), as a set of more than a hundred bidirectional measures (called bands), of the same region (called ground truth map: GT), needs reduction dimensionality. Indeed the bands don’t all contain the information; some bands are irrelevant like those affected by various atmospheric effects, see Figure.3, and decrease the classification accuracy. Finaly there exist redundant bands, must be avoided. We can reduce the dimensionality of hyperspectral images by selecting only the relevant bands (feature selection or subset selection methodology), or extracting, from the original bands, new bands containing the maximal information about the classes, using any functions, logical or numerical (feature extraction methodology) [8][9][11] . Here we introduce an algorithm based on mutual information, reducing dimensionality in too steps: pick up the relevant bands first, and avoiding redundancy second. We illustrate the principea of this algorithm using synthetic bands for the scene of HIS AVIRIS 92AV3C [1] , Figure.1.Then we approve its effectiveness with applying it to real datat of HSI AVIRIS 92AV3C. So each pixel is shown as a vector of  220 components. Figure.2. shows the vector pixels notion [7 ]. So reducing dimensionality means selecting only the dimensions caring a lot of information regarding the classes.

The Hyperspectral image AVIRIS 92AV3C (Airborne Visible Infrared Imaging Spectrometer)[2] contains 220 images taken on the region "Indiana Pine" at "north-western Indiana", USA [1]. The 220 called bands are taken between \unit{0.4}{\micro}m and \unit{2.5}{\micro}m. Each band has 145 lines and 145 columns. The ground truth map is also provided, but only 10366 pixels (49\%) are labeled fro 1 to 16. Each label indicates one from 16 classes. The zeros indicate pixels how are not classified yet, Figure.1.
The hyperspectral image AVIRIS 92AV3C contains numbers between 955 and 9406. Each pixel of the ground truth map has a set of 220 numbers (measures) along the hyperspectral image. Those numbers (measures) represent the reflectance of the pixel in each band. So the pixel is shown as vector off 220 components. Figure .2.

\begin{figure}[!th]
\centering
\includegraphics[width=3.5in]{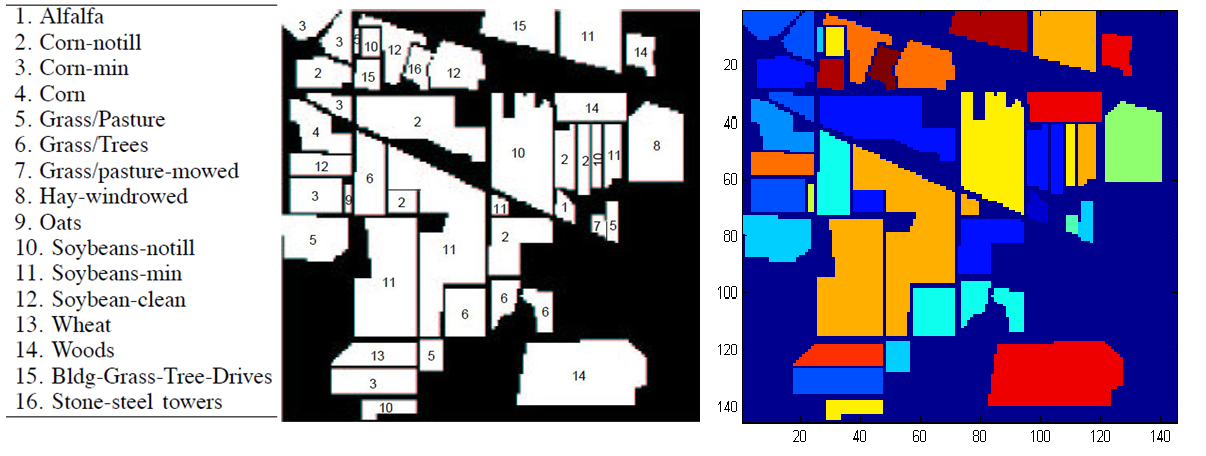}
\caption{The Ground Truth map of AVIRIS 92AV3C and the 16 classes }
\label{fig_sim}
\end{figure}

We can also note that not all classes are carrier of information. In Figure.5, for example, we can show the effects of atmospheric effects on bands: 155, 220 and other bands. This Hyperspectral Image presents the problematic of dimensionality reduction.

Figure.2 shows the vector pixel’s notion [7]. So reducing dimensionality means selecting only the dimensions caring a lot of information regarding the classes.
\begin{figure}[!h]
\centering
\includegraphics[width=3.5in]{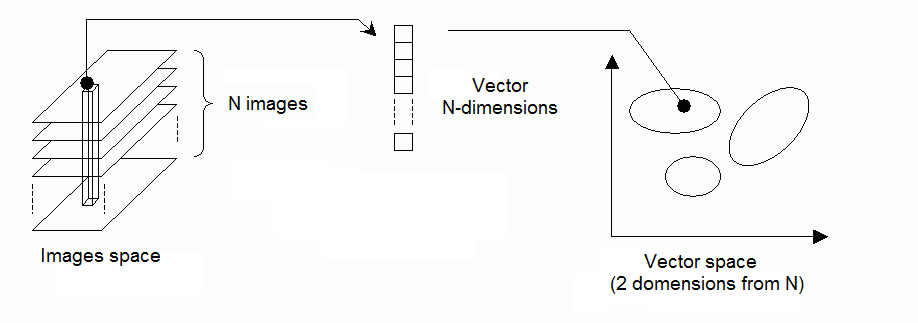}
\caption{The notion of  pixel vector }
\end{figure}

\section{Mutual Information Based Feature Selection}

 

\subsection{Definition of Mutual Information}
This is a measure of exchanged information between tow ensembles of random variables A and B:
\[
I(A,B)=\sum\;p(A,B)\;log_2\;\frac{p(A,B)}{p(A).p(B)}
\]

Considering the ground truth map, and bands as ensembles of random variables, we calculate their interdependence. Fano [14] has demonstrated that as soon as mutual information of already selected feature has high value, the error probability of classification is decreasing, according to the formula bellow:
\[\;\frac{H(C/X)-1}{Log_2(N_c)}\leq\;P_e\leq\frac{H(C/X)}{Log_2}\; \]with :
\[
\;\frac{H(C/X)-1}{Log_2(N_c)}=\frac{H(C)-I(C;X)-1}{Log_2(N_c)}\; \] and :
\[    P_e\leq\frac{H(C)-I(C;X)}{Log_2}=\frac{H(C/X)}{Log_2}\; \]
The expression of conditional entropy $H(C/X)$ is calculated between the ground truth map (i.e. the classes $C$) and the subset of bands candidates $X$. $ N_{c}$ is the number of classes. So when the features $X$ have a higher value of mutual information with the ground truth map, (is more near to the ground truth map), the error probability will be lower. But its difficult to compute this conjoint mutual information $I(C;X)$, regarding the high dimensionality [14].\\
Figure.4. shows the MI between the GT and synthetic bands. The figure .6 shows the MI between the GT and the real bands of HIS AVIRIS 92AV3C [1].\\
Many studies use a threshold to choice the relevant bands. Guo [3] uses the mutual information to select the top ranking band, and a filter based algorithm to decide if there neighbours are redundant or not. Sarhrouni et al. [17] use also a filter strategy based algorithm on MI to select bands. A wrapper strategy based algorithm on MI, Sarhrouni et al. [18] is also introduced.\\
By a thresholding, for example with a threshold 0.4, see Figure.5, we eliminate the no informative bands: $A_3$, $A_7$ and $A_9$. With other threshold, we can retain fewer bands. We can visually verify this effectiveness of MI to choice relevant features in Figure.4.

\subsection{Symmetric Uncertainty}
This is one of normalized form of Mutual Information; introduced by Witten and Frank, 2005 [19]. It’s defined as bellow: 
\[
U(A,B)=2.\frac{MI(A,B)}{H(A)+H(B)}
\]
$H(X)$ is the Entropy of set random variable $X$. Some studies use this $U$ for recalling images in medical images treatment [9]. Numerous studies use Normalized Mutual Information [20][21][22].\\
Figure.3 shows that $symmetric$ $uncertainty$ means how much information is partaged between $A$ and $B$ relatively at all information contained in both $A$ and $B$
\begin{figure}[!th]
\centering
\includegraphics[width=3.0in]{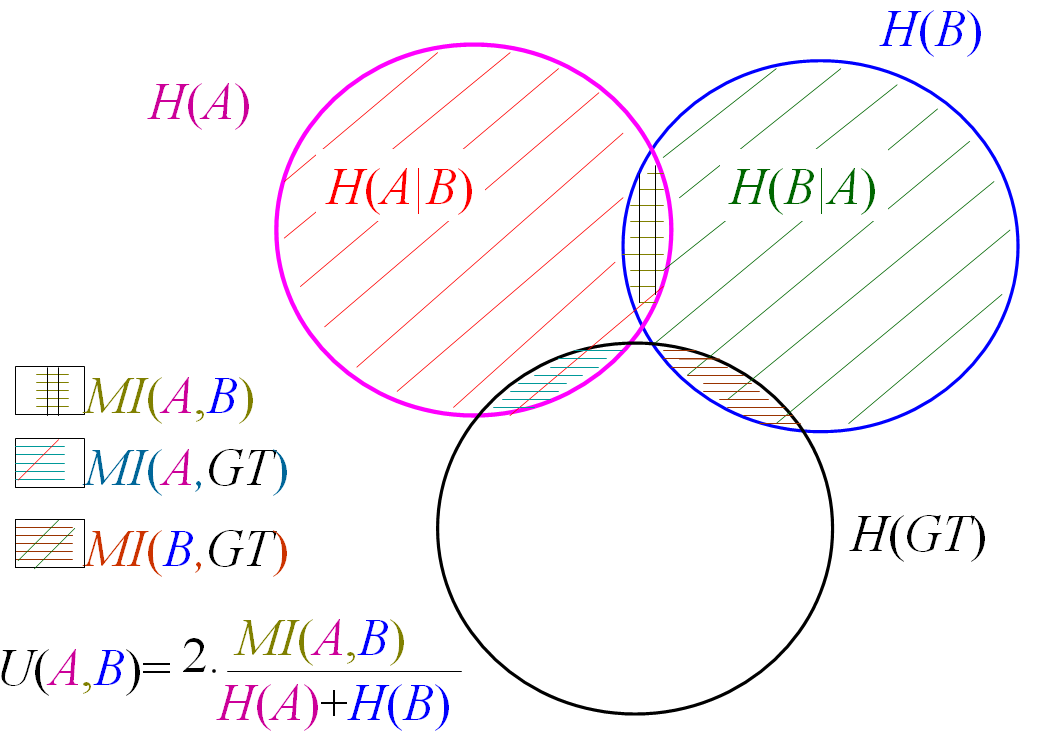} 
\caption{Illustration of Symetric Uncertainty}
\end{figure}

\section{Principe of the Proposed Method and Algorithm}
For this section we synthesize 19 bands from the GT, Figure.1, by adding noise, cutting some substances etc. see Figure.4. Each band has 145 lines and 145 columns. The ground truth map is also provided, but only 10366 pixels are labelled from 1 to 16. Each label indicates one from 16 classes. The zeros indicate pixels how are not classified yet, Figure.2. We can show the Mutual information of GT and the synthetic bands at Figure.5.
\begin{figure}[!th]
\centering
\includegraphics[width=3.50in]{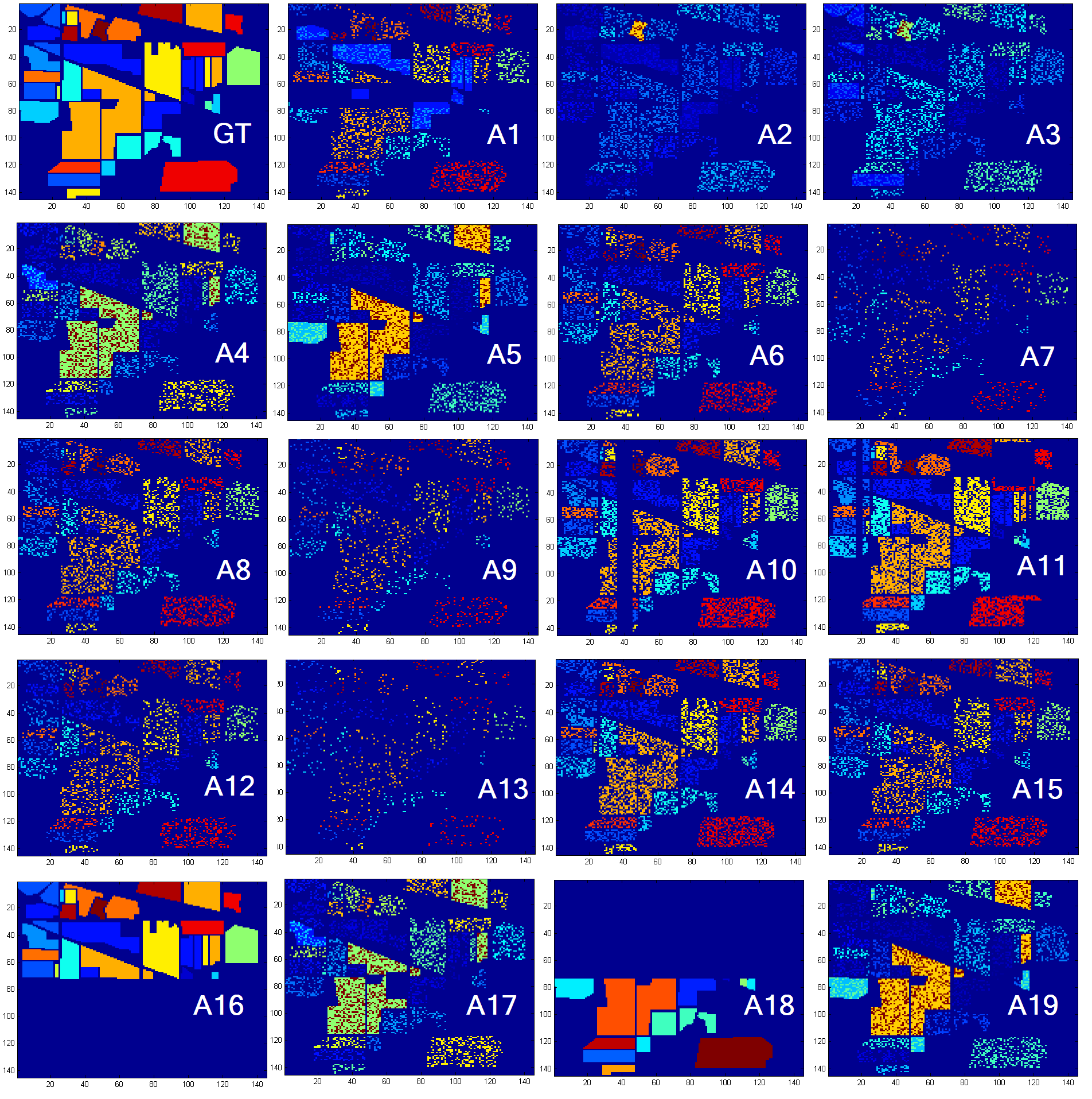}
\caption{The synthetic bands used for the study .}
\end{figure}

\begin{figure}[!th]
\centering
\includegraphics[width=3.5in]{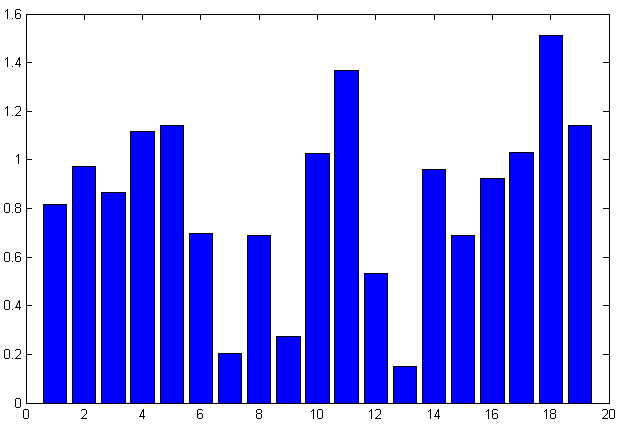}
\caption{Mutual Information of GT and synthetic bands   .}
\end{figure}

\subsection{Principe to select relevant bands}
With a threshold 0.4 of MI calculated in Figure.4 we obtain 16 relevant bands $ A_i$:\\
 with      i=\{1,2,3,4,5,6,8,10,11,12,14,15,16,17,18,19\}.\\
 We can visually verify the resemblance of GT and the bands more informative, bout in synthetic and the real data bands of AVIRIS 92AV3C. See Figure.6.

\subsection{Principe of  no Redundant Bands Detection}
First: We order the remaining bands, in increasing order of there MI with the GT. So we have:\\
$\{A_{12} A_8 A_{15 }A_6 A_1 A_3 A_{14 }A_{16} A_2 A_{10 }A_{17 }A_4 A_{19} A_5 A_ {11} A_ {18}\}$\\

Second: We fixe a threshold to control redundancy, here 0.7. Then we compute the Symmetric Uncertainty: $U (A_{i},A_{j})$ for all couple $( i, j )$ of the ensemble:\\
 $S=\{8,15,6,1,3,14,16,2,10,17,4,19,5,11,18\}$. \\

Observation 1 : Figure.4 shows that the band $A_{17}$ is practically the same at  $A_{4}$.  Table I shows  $U (A_{17}, A_{4})$ near to 100\% (0.95). So this indicates a high redundancy.\\

Observation 2 : Figure.4 shows that the bands $ A_{16}$ and $A_{18}$ are practically disjoint, i.e. they are not redundant. Table.I. shows $U(A_{16}, A_{18})$ =0.07. So this indicates no redundancy. So the ensemble of selected bands became $SS= \{16, 18\}$. $A_{16}, A_{18}$ will be discarded from the Table .I. Algorithm 1 shows more details.\\
 Now we can emit this rule:
 
\textit{Rule: Each band candidate will be added at SS if and only if their Symmetric Uncertainty values with all elements off SS, are less than the thresholds (here 0.7).}\\

Algorithm 1 shows more detailsimplements this rule.

\begin{table}[!h]
\center
\caption{THE SYMMETRIC UNCERTAINTY OF THE RELEVANT SYNTHETIC BANDS. }
\includegraphics[width=5.50in]{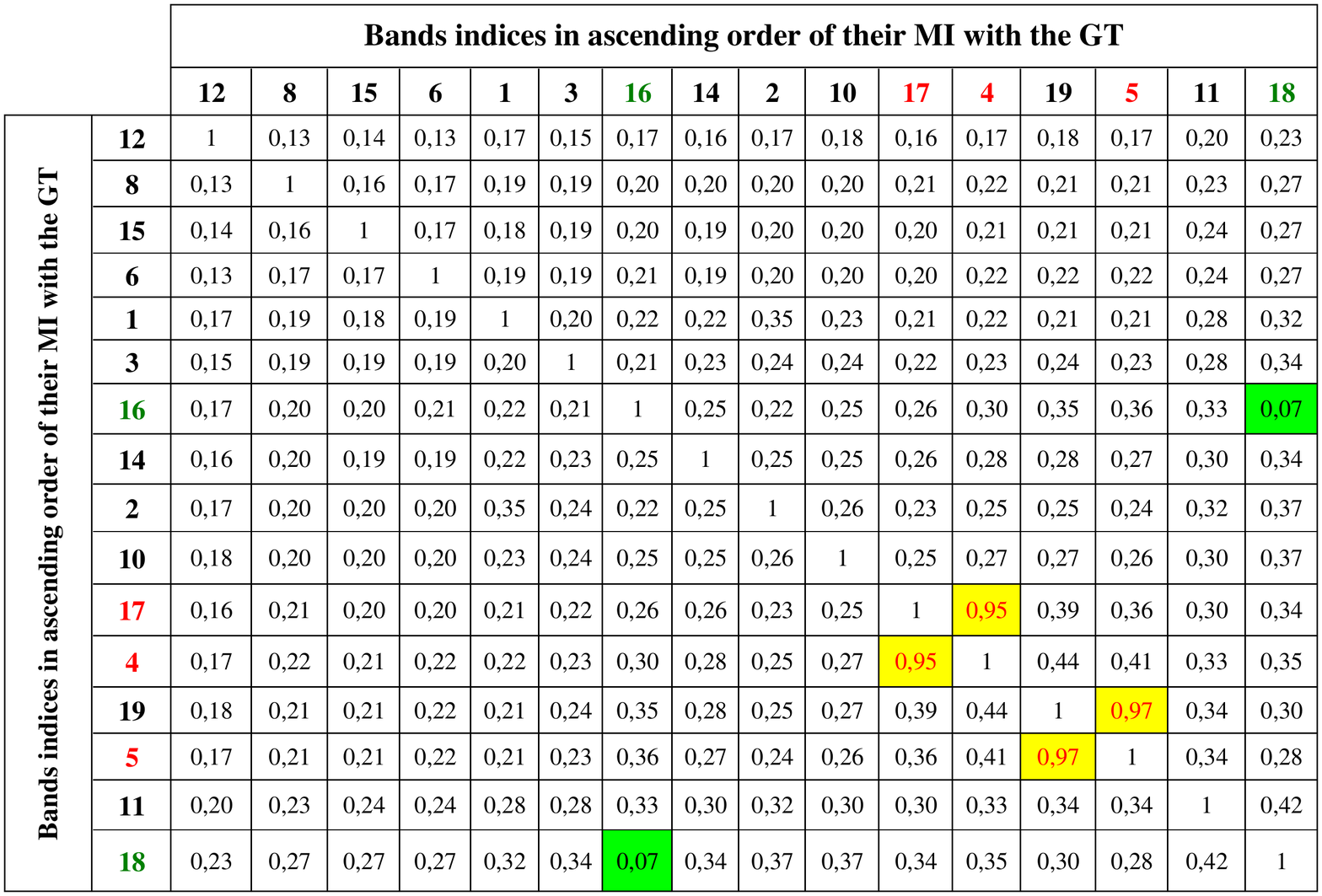}
\end{table}

\begin{algorithm}  
\vspace{0.20cm}                    
\caption{: $Band$ is the HSI. Let $Th$$_{relevance}$ the threshold for selecting bands more informative, $Th$$_{redundancy}$  the threshold for redundancy control. }  
\center        
\label{algo 1}  
\begin{algorithmic}
\STATE 1) Compute the Mutual Information $(MI)$ of the bands and the Ground Truth map.
\STATE 2) Make bands in ascending order by there $MI$ value 
\STATE 3) Cut the bands that have a lower value than the threshold $Th$$_{relevance}$, the subset remaining is $S$.
\STATE 4) Initialization:  $n=length(S), i=1$,  $D$ is a bidirectional array values=1;\\
//any value greater than 1 can be used, it's useful in step 6)
\STATE 5) Computation of bidirectional Data $D(n,n)$:

\FOR{  1:=1  to n step 1 }
\FOR{	 j:=i+1  to n step 1 }

\STATE $D(i,j)= U(Band_{S(i)},Band_{S(j)});$  
\STATE	 // with $U(A,B)=\frac{MI(A,B)}{H(A)+H(B)}$
\ENDFOR
\ENDFOR
\STATE	  //Initialization of the Output of the algorithm
\STATE 6) $SS=\{\}$ ;
\WHILE{$ min(D)<Th_{redundancy}$}
\STATE 	// Pick up the argument of the minimum of D
\STATE $(x,y)= argmin(D(.,.));$     
\IF {$ \forall \  l \in SS\, \  D(x,l)<Th_{redundancy}$}  
\STATE	 // $x$ is not redundant with the already selected bands
 \STATE $SS=SS \cup \{x\}$   
\ENDIF
\IF {$ \forall \  l \in SS\, \  D(y,l)<Th_{redundancy}$}  
\STATE	 // $y$ is not redundant with the already selected bands
 \STATE $SS=SS \cup \{y\}$   
\ENDIF
\STATE $D(x,y)=1$; $D(x,y)=1$;	
// The cells $D(x,y)$  and $D(y,x)$ will not be checked as minimum again	
\ENDWHILE

\STATE 7)  Finish: The  final subset  SS  contains bands according to the the couple of  thresholds \ ( $Th_{relevance}$,$T_{redundancy}$). 

\end{algorithmic}
\end{algorithm}

\section{Application On HIS AVIRIS 92AV3C  }
The Agorithm.1 implement the proposed method .

We apply the proposed algorithm on the hyperspectral image AVIRIS 92AV3C [1], 50\% of the labelled pixels are randomly chosen and used in training; and the other 50\% are used for testing classification [3][17][18]. The classifier used is the SVM [5][12] [4].

\subsection{Mutual Information Curve of Bands}
From the remaining subset bands, we must eliminate no informative ones, bay thresholding, see the proposed algorithm. Figure.6 gives the MI of the HSI AVIRIS 92AV3C with the ground truth GT.

\begin{figure}[!h]
\centering
\includegraphics[width=3.5in]{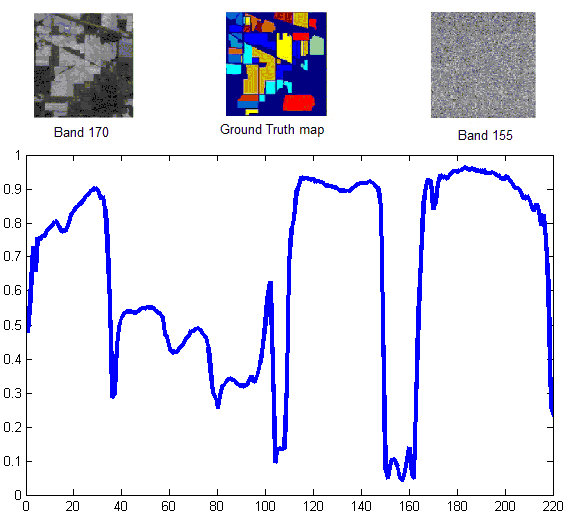}
\caption{Mutual information of GT and AVIRIS bands.}
\end{figure}

\subsection{Results}
From the remaining subset bands, we must eliminate redundant ones using the proposed algorithm. Table II gives the accuracy off classification for a number of bands with several thresholds.

\subsection{Discussion}

Results in Table.II allow us to distinguish six zones of couple values of thresholds $(TH,IM)$:\\

Zone1: This is practically no control of relevance and no control of redundancy. So there is no action of the algorithm.\\

Zone2: This is a hard selection: a few more relevant and no redundant bands are selected. \\

Zone3: This is an interesting zone. We can have easily 80\% of classification accuracy with about 40 bands.\\

Zone4: This is the very important zone; we have the very useful behaviours of the algorithm. For example with a few numbers of bands 19 we have classification accuracy 80\%.\\

Zone5: Here we make a hard control of redundancy, but the bands candidates are more near to the GT, and they my be  more redundant. So we can’t have interesting results.\\ 

Zone6: When we do not control properly the relevance, some bands affected bay transfer affects may be non redundant, and can be selected, so the accuracy of classification is decreasing. \\

 Partial conclusion: This algorithm is very effectiveness for redundancy and relevance control, in feature selection area. \\

 The most difference of this algorithm regarding previous works is the separation of the tow process: avoiding redundancy and selecting more informative bands. Sarhrouni et al. [17] use also a filter strategy based algorithm on MI to select bands, and an another wrapper strategy algorithm also based on MI [18], Guo[3] used a filter strategy with threshold control redundancy, but in those works, the tow process, i.e avoiding redundancy and avoiding no informative bands, are made at the same time by the same threshold.\\
\begin{table*}[!th]
\center
\caption{Classification Accuracy for several couples of thresholds (TH,IM) and their corresponding number of bands retained. }
\includegraphics[width=5.8in]{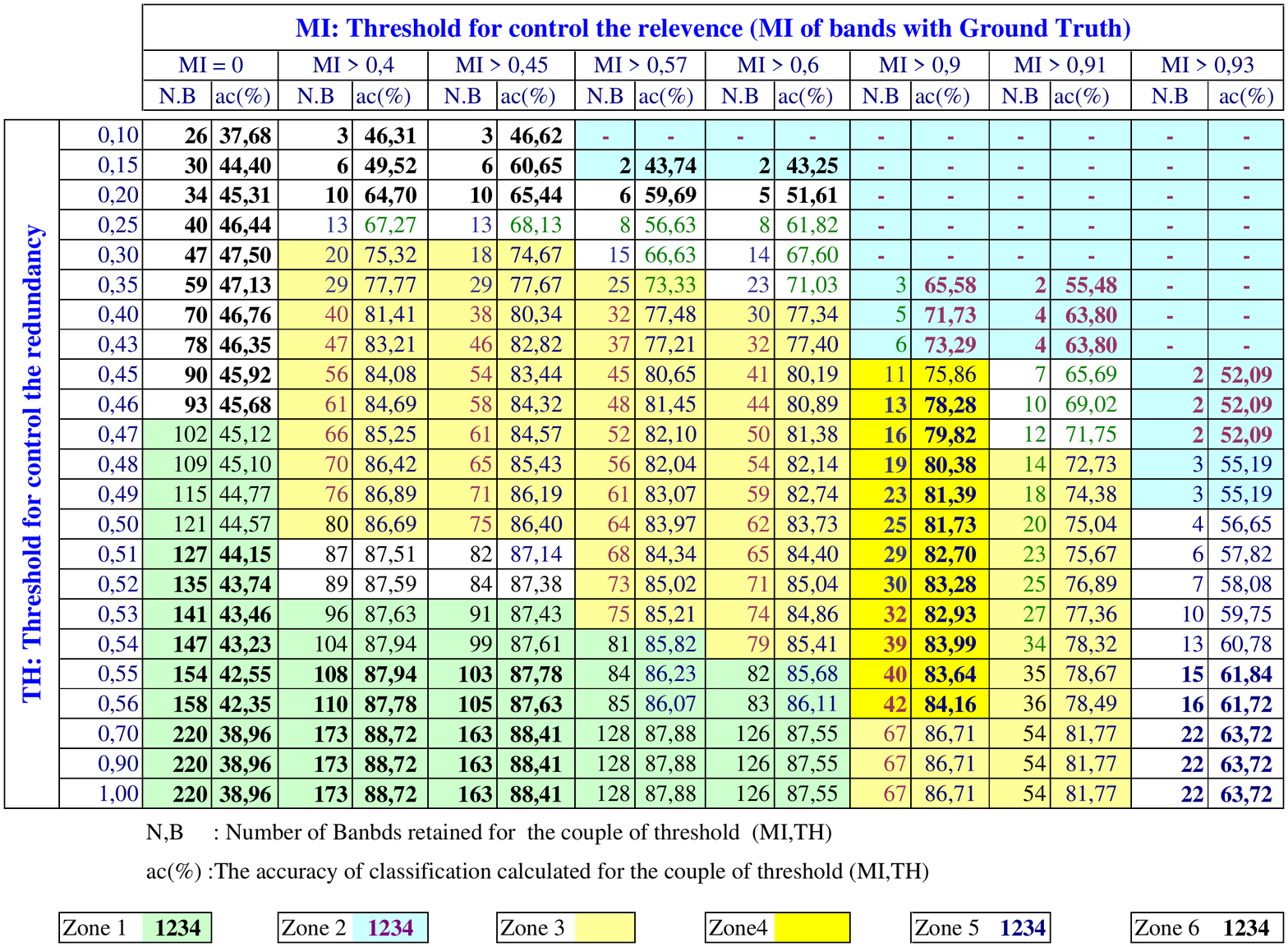}
\end{table*}
\begin{figure*}[!th]
\centering
\includegraphics[width=5.50in]{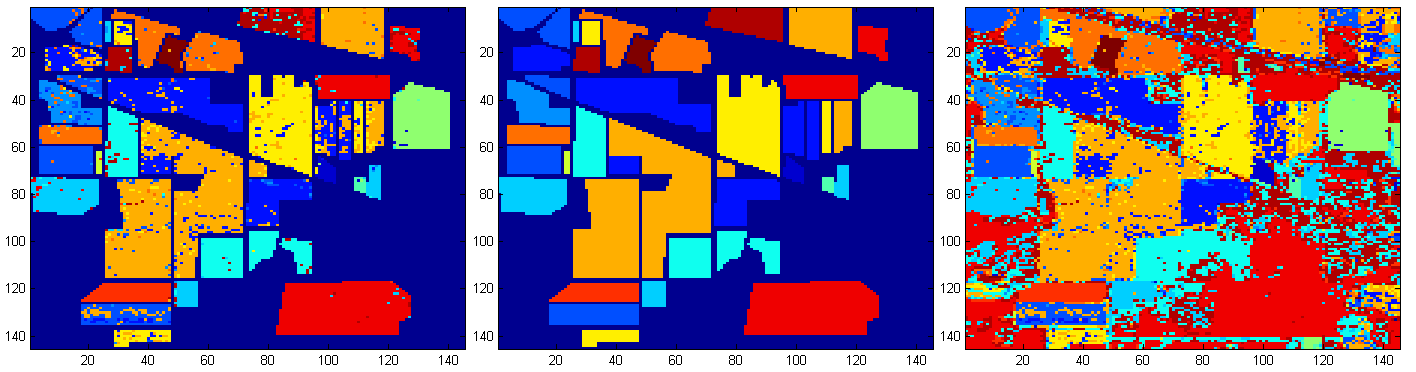} 
\caption{In the middle the GT of AVIRIS 92AV3C. In the left: Reconstructed Truth map (GT) with the proposed algorithm for TH=0.56 and MI=0.9; the  accuracy = 84.16 \% for only 42 bands. In right the generalization of classification for all Indiana Pine regions.}
\end{figure*}

Figure.7. illustrates the reconstruction of the ground truth map GT, for a redundancy threshold 0.56 and relevance threshold IM=0.9. The accuracy classification is 84.16\% for 42 bands selected. The figure.7 gives also a general classification of the entire scene Indiana Pin [1]; the pixels not labelled in GT, are here classified. This illustrates the power of generalisation of the proposed method.\\

We can not here that Hui Wang [15] uses two axioms to characterize feature selection. Sufficiency axiom: the subset selected feature must be able to reproduce the training simples without losing information. The necessity axiom "simplest among different alternatives is preferred for prediction". In the algorithm proposed, reducing error uncertainty between the truth map and the estimated minimize the information loosed for the samples training and also the predicate ones.
We not also that we can use the number of features selected like condition to stop the search. [16].

\section{Conclusion}
Until recently, in the data mining field, and features selection in high dimensionality the problematic is always open. Some heuristic methods and algorithms have to select relevant and no redundant subset features. In this paper we introduce an algorithm in order to process separately the relevance and the redundancy. We apply our method to classify the region Indiana Pin with the Hyperspectral Image AVIRIS 92AV3C. This algorithm is a Filter strategy (i.e. with no call to classifier during the selection). In the first step we use mutual information to pick up relevant bands by thresholding (like most method already used). The second step introduces a new algorithm to measure redundancy with Symmetric Uncertainty coefficient. We conclude the effectiveness of our method and algorithm the select the relevant and no redundant bands. This algorithm allows us a wide area of possible fasted applications. But the question is always open: no guaranties that the chosen bands are the optimal ones; because some redundancy can be important to reinforcement of learning classification system. So the thresholds controlling relevance redundancy is a very useful tool to calibre the selection, in real time applications. This is a very positive point for our algorithm; it can be implemented in a real time application, because in commercial applications, the inexpensive filtering algorithms are urgently preferred.


\begin{thebibliography}{1}
\bibitem{1}
D. Landgrebe, “On information extraction principles for hyperspectral data: A white paper,” Purdue University, West Lafayette, IN, Technical Report, School of Electrical and Computer Engineering, 1997. Téléchargeable ici :
  http://dynamo.ecn.purdue.edu/~landgreb/whitepaper.pdf.

\bibitem{2}
ftp://ftp.ecn.purdue.edu/biehl/MultiSpec/
\bibitem{3}
Baofeng Guo, Steve R. Gunn, R. I. Damper Senior Member, "Band Selection for Hyperspectral Image Classification Using Mutual Information" , IEEE and J. D. B. Nelson. IEEE GEOSCIENCE AND REMOTE SINSING LETTERS, Vol .3, NO .4, OCTOBER 2006.
\bibitem{4}
Baofeng Guo, Steve R. Gunn, R. I. Damper, Senior Member, IEEE, and James D. B. Nelson."Customizing Kernel Functions for SVM-Based Hyperspectral Image Classification",  IEEE TRANSACTIONS ON IMAGE PROCESSING, VOL. 17, NO. 4, APRIL 2008.
\bibitem{5}
Chih-Chung Chang and Chih-Jen Lin, LIBSVM: a library for support vector machines. ACM Transactions on Intelligent Systems and Technology , 2:27:1--27:27, 2011. Software available at http://www.csie.ntu.edu.tw/~cjlin/libsvm.
\bibitem{6}
Nathalie GORRETTA-MONTEIRO , Proposition d'une approche de'segmentation d’images hyperspectrales. PhD thesis. Universite Montpellier II. Février 2009.
\bibitem{7}
David Kernéis." Amélioration de la classification automatique des fonds marins par la fusion multicapteurs acoustiques". Thèse,  ENST BRETAGNE, université de Rennes. Chapitre3, Réduction de dimensionalité et classification,Page.48. Avril 2007.
\bibitem{8}
Kwak, N and Choi, C. "Featutre extraction based on direct calculation of mutual information".IJPRAI VOL. 21, NO. 7, PP. 1213-1231, NOV. 2007 (2007).
\bibitem{9}
 Nojun Kwak and C. Kim,"Dimensionality Reduction Based on ICA Regression Problem". ARTIFICIAL NEURAL NETWORKS-ICANN 2006. Lecture Notes in Computer Science, 2006, Isbn 978-3-540-38625-4,  Volume 1431/2006.
\bibitem{10}
 Huges, G. Information Thaory,"On the mean accuracy of statistical pattern recognizers". IEEE Transactionon Jan 1968, Volume 14, Issue:1, p:55-63, ISSN 0018-9448 DOI: 10.1109/TIT.1968.1054102.

\bibitem{11}
YANG, Yiming, and Jan O. PEDERSEN, 1997.A comparative study of feature selection in text categorization. In: ICML. 97: Proceedings of the Fourteenth International Conference on Machine Learning. San Francisco, CA, USA:Morgan Kaufmann Publishers Inc., pp. 412.420.
\bibitem{12}
Chih-Wei Hsu; Chih-Jen Lin,"A comparison of methods for multiclass support vector machines" ;Dept. of Comput. Sci.  Inf. Eng., Nat Taiwan Univ. Taipei Mar 2002, Volume: 13 I:2;pages: 415 - 425 ISSN: 1045-9227, IAN: 7224559, DOI: 10.1109/72.991427
\bibitem{13}
Bermejo, P.; Gamez, J.A.; Puerta, J.M."Incremental Wrapper-based subset Selection with replacement: An advantageous alternative to sequential forward selection" ;Comput. Syst. Dept., Univ. de Castilla-La Mancha, Albacete; Computational Intelligence and Data Mining, 2009. CIDM '09. IEEE Symposium on March 30 2009-April 2 2009, pages: 367 - 374, ISBN: 978-1-4244-2765-9, IAN: 10647089, DOI: 10.1109/CIDM.2009.4938673
\bibitem{14}
 Lei Yu, Huan Liu,"Efficient Feature Selection via Analysis of Relevance and Redundancy", Department of Computer Science and Engineering; Arizona State University, Tempe, AZ 85287-8809, USA, Journal of Machine Learning Research 5 (2004) 1205-1224.
\bibitem{15}
Hui Wang, David Bell, and Fionn Murtagh, "Feature subset selection based on relevance" , Vistas in Astronomy, Volume 41, Issue 3, 1997, Pages 387-396.
\bibitem{16}
P. Bermejo, J.A. Gámez, and J.M. Puerta, "A GRASP algorithm for fast hybrid (filter-wrapper) feature subset selection in high
dimensional datasets", presented at Pattern Recognition Letters, 2011, pp.701-711.
\bibitem{17}
E. Sarhrouni, A. Hammouch, D. Aboutajdine "Dimensionality reduction and classification feature using mutual information applied to hyperspectral images: a filter strategy based algorithm". Appl. Math. Sci., Vol. 6, 2012, no. 101-104, 5085-5095.\bibitem{18}
E. Sarhrouni, A. Hammouch, D. Aboutajdine " Dimensionality reduction and classification feature using mutual information applied to hyperspectral images: a wrapper strategy algorithm based on minimizing the error probability using the inequality of Fano". Appl. Math. Sci., Vol. 6, 2012, no. 101-104, 5073-5084.
\bibitem{19}
Ian H. Witten and Eibe Frank. Data Mining: Practical Machine Learning Tools and Techniques. Morgan Kaufmann, San Francisco, 2 edition, 2005.
\bibitem{20}
E. R. E. Denton, M. Holden, E. Christ, J. M. Jarosz, D. Russell-Jones, J. Goodey, T. C. S. Cox, and D. L. G. Hill, “The identification of cerebral volume changes in treated growth hormonedeficient adults using serial 3D MR image processing,” Journalof Computer Assisted Tomography, vol. 24, no. 1, pp. 139 - 145, 2000.
\bibitem{21}
M. Holden, E. R. E. Denton, J. M. Jarosz, T. C. S. Cox, C. Studholme, D. J. Hawkes, and D. L. G. Hill, “Detecting small anatomical changes with 3D serial MR subtraction images,” in Medical Imaging: Image Processing, K. M. Hanson, Ed., SPIE Press, Bellingham, WA, 1999, vol. 3661 of Proc. SPIE, pp. 44 - 55.
\bibitem{22}
M. Otte, “Elastic registration of fMRI data using B´ezier-spline transformations,” IEEE Transactions on Medical Imaging, vol. 20, no. 3, pp. 193–206, 2001.
\end{thebibliography}
\end{document}